\documentclass[letterpaper, 10 pt, conference]{ieeeconf}  % Comment this line out if you need a4paper
\usepackage{graphicx}
\usepackage{amsmath}
\usepackage{amssymb}
\usepackage{booktabs}
\let\labelindent\relax
\usepackage{enumitem}
\usepackage{multirow}
\usepackage{xcolor}
\usepackage[
backend=biber,
style=numeric,
sorting=none
]{biblatex}
\usepackage{hyperref} 
\IEEEoverridecommandlockouts                              % This command is only needed if 
\title{\huge
\textbf{\textit{ContextualFusion}}: Context-Based Multi-Sensor Fusion for 3D Object Detection in Adverse Operating Conditions
}

\author{Shounak Sural$^{1}$, Nishad Sahu$^{1}$ and Ragunathan (Raj) Rajkumar$^{1}$% <-this % stops a space
\thanks{$^{1}$The authors are associated with the Department of Electrical and Computer Engineering at Carnegie Mellon University, Pittsburgh, PA, 15213, USA.
        Email: {\tt\small \{ssural,nsahu,rajkumar\}@andrew.cmu.edu}}%
}
\usepackage{tikz}
\newcommand\copyrighttext{%
  \footnotesize \textcopyright 2024 IEEE. Personal use of this material is permitted.
  Permission from IEEE must be obtained for all other uses, in any current or future
  media, including reprinting/republishing this material for advertising or promotional
  purposes, creating new collective works, for resale or redistribution to servers or
  lists, or reuse of any copyrighted component of this work in other works.}
\newcommand\copyrightnotice{%
\begin{tikzpicture}[remember picture,overlay]
\node[anchor=south,yshift=10pt] at (current page.south) {\fbox{\parbox{\dimexpr\textwidth-\fboxsep-\fboxrule\relax}{\copyrighttext}}};
\end{tikzpicture}%
}

\begin{document}

\maketitle
\copyrightnotice
\thispagestyle{empty}
\pagestyle{empty}

\begin{abstract}
The fusion of multimodal sensor data streams such as camera images and lidar point clouds plays an important role in the operation of autonomous vehicles (AVs). Robust perception across a range of adverse weather and lighting conditions is specifically required for AVs to be deployed widely. While multi-sensor fusion networks have been previously developed for perception in sunny and clear weather conditions, these methods show a significant degradation in performance under night-time and poor weather conditions. In this paper, we propose a simple yet effective technique called \textit{ContextualFusion} to incorporate the domain knowledge about cameras and lidars behaving differently across lighting and weather variations into 3D object detection models. Specifically, we design a Gated Convolutional Fusion (\textit{GatedConv}) approach for the fusion of sensor streams based on the operational context. To aid in our evaluation, we use the open-source simulator CARLA to create a multimodal adverse-condition dataset called \textit{AdverseOp3D} to address the shortcomings of existing datasets being biased towards daytime and good-weather conditions. Our \textit{ContextualFusion} approach yields an mAP improvement of $\textbf{6.2\%}$ over state-of-the-art methods on our context-balanced synthetic dataset. Finally, our method enhances state-of-the-art 3D objection performance at night on the real-world NuScenes dataset with a significant mAP improvement of $\textbf{11.7\%}$.


%Moreover, on the entire NuScenes dataset, we get an improvement of $0.91\%$ for overall mAP with over $3\%$ AP improvement for the classes- motorcycles and bicycles. 

\end{abstract}

%%%%%%%%% BODY TEXT

\section{Introduction}\label{introduction}
 Autonomous vehicles (AVs) are intended to drive by themselves in real-world scenarios with little or no human involvement. These AVs are often equipped with multiple sensors including cameras, lidars, radars, ultrasonic sensors, GPS and IMU. Research in the AV domain has gained a lot of popularity because of its promise of making transportation safer. Companies like Waymo and GM Cruise have started deploying autonomous robotaxi fleets in limited areas, primarily in good weather conditions. 
 
 Automotive crashes cause fatalities, injuries and road blockages. Most of these crashes are due to human error, and removing humans from all or major portions of driving can potentially avoid or minimize a large number of crashes. To meet this goal, automated driving systems, however, must perceive their operational environment reliably. A fully-autonomous AV must be able to see 360 degrees in the horizontal field of view, and about $\pm$ 15-20 degrees in the vertical field of view without any blind spots. 
 \begin{figure}
\centering
\includegraphics[scale=0.065]{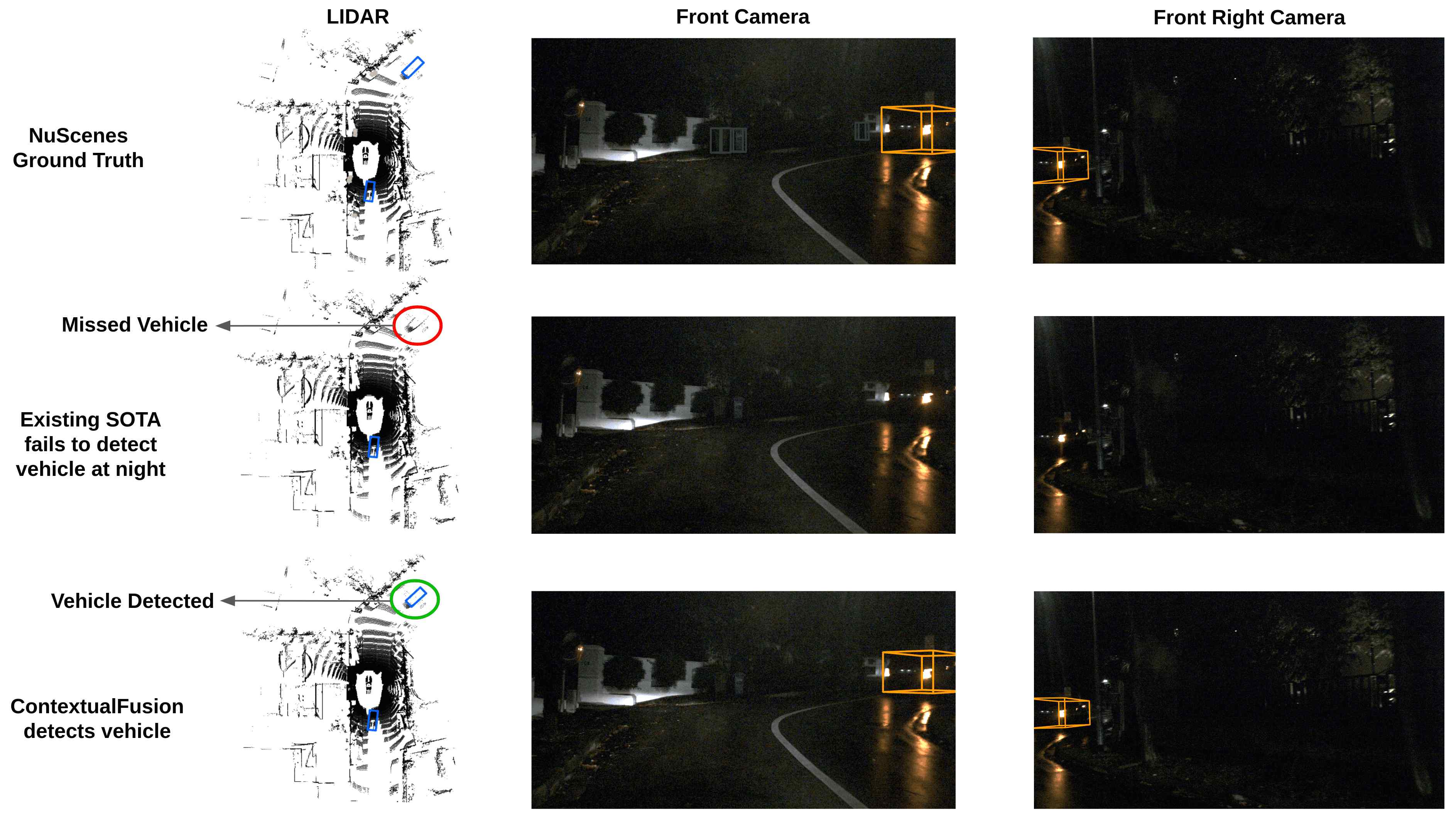}
\caption{Comparison of NuScenes \cite{nuscenes} ground truth (top row) with predictions from existing state-of-the-art models (middle row) vs our \textit{ContextualFusion} predictions (bottom row) on night-time NuScenes data} 
\label{fig:comparison_2img}
\vspace{-5mm}
\end{figure} 
 A combination of camera, lidar and radar sensors provides significant redundancy towards achieving this goal. In this paper, we focus on the fusion of camera images and lidar point clouds in adverse conditions. Cameras provide rich pixel-level information but can be blinded by direct sunlight and experience significant performance drop in night-time and low-lighting conditions. Lidars, being active sensors providing 3D information, work equally well across both day- and night-time conditions. However, lidar beams can also be scattered by heavy rain and fog leading to false positives in object detection. Recent studies have shown that fusing sensor data using multimodal deep-learning models is often effective in improving perception performance over the usage of individual sensor-based predictions \cite{bijelic2020seeing,bevfusion,transfusion}.  \textit{BEVFusion} \cite{bevfusion} is one such baseline that is the current state of the art on the NuScenes \cite{nuscenes} dataset and provides inference outputs in a reasonable amount of time. However, many existing approaches are context-agnostic and are trained on data predominantly in good weather and lighting conditions. This causes a drastic degradation in performance when these models are inferenced in night-time and rainy conditions, which are needed for real-world deployment. In this paper, we explore the usage of a context-aware approach to sensor fusion and show that it can significantly improve perception performance.
 
For instance, Figure \ref{fig:comparison_2img} illustrates the typical performance of the state-of-the-art (SOTA) BEVFusion method in comparison to our approach for a scenario at night in a dimly-lit environment from the NuScenes dataset, where a car in front of the ego vehicle is around a curve on the road. The blue boxes in the lidar image and yellow boxes in the camera image indicate the corresponding detected boxes. These yellow boxes are not seen with BEVFusion, showing that it is ineffective in such cases, failing to detect the vehicle in both the lidar and camera viewpoints. Such failures can be attributed to the over-reliance of BEVFusion on the camera sensor uniformly across day- and night-time conditions, which results in a big drop in night-time performance. Our {\textit{ContextualFusion}} approach handles all such scenarios by intelligently assigning appropriate importance to lidar and camera modalities across varying lighting and rain conditions. This context-based approach results in an accurate detection of the vehicle as seen in both lidar and camera views.

In summary, the contributions of our work are as follows:

\begin{enumerate}
    \item We propose a research direction for intelligent sensor fusion, based on the operating context of AVs.
    \item We design a day/night and clear/rainy weather information-based fusion model that incorporates domain knowledge about the limitations of camera and lidar sensors into a deep learning framework. 
    \item We design a simple yet effective context-based gating and convolutional fusion module to differentially weigh camera and lidar features during fusion.
    \item We create a public multimodal dataset in the NuScenes format called \textit{AdverseOp3D} that uses the open-source CARLA simulator to extensively simulate adverse operating conditions. 
\end{enumerate}

The rest of this paper is organized as follows. Section \ref{sec:related-works} discusses related work in this domain. Section \ref{sec:methodology} describes {\textit{ContextualFusion}} for 3D object detection and our synthetic data generation process. Section \ref{sec:experiments} evaluates the performance of {\textit{ContextualFusion}} and compares it to BEVFusion and other similar techniques. Finally, Section \ref{sec:conclusion} concludes this paper and also provides directions for future research. 

%[Maybe subsections currently in introduction can partially be moved to related work? @Nishad]

%(Shounak will write the below three things here)
%[Research questions]

%[List of contributions]

%[Organization of sections]
Our code and the \textit{AdverseOp3D} dataset are publicly released at \href{https://github.com/ssuralcmu/ContextualFusion.git}{https://github.com/ssuralcmu/ContextualFusion.git}. 

\section{Related Work}
\label{sec:related-works}
In this section, we describe some related work. Specifically, we discuss related studies on 3D object detection, perception in adverse weather and the usage of AV simulators.
\subsection{3D Object Detection}
\label{3dod}
3D object detection has several applications in the fields of autonomous driving and robotics. It refers to detecting object class, location, orientation and dimensions in 3D space. Many early methods for 3D object detection, such as \cite{schneiderman2000statistical,guil1999planar}, relied on statistical methods, geometric priors and handcrafted techniques. After the development of faster and cheaper hardware platforms, Deep Neural Networks (DNNs) became popular for 3D object detection applications. Today, many DNN-based networks \cite{bevfusion,yin2021center,chen2017multi,wu2022transformation} provide high accuracy on datasets like NuScenes\cite{nuscenes} and KITTI \cite{geiger2012we} using camera and lidar sensor modalities for 3D object detection. Vision-radar models have also been extensively investigated \cite{vision_radar}. Although these DNN models work well for 3D-object detection in clear lighting conditions, their performance drops significantly at low-visibility conditions even with state-of-the-art networks. The performance of these methods is also affected under rain, fog and snow conditions. Thus, more robust models are required to obtain better performance in such inclement conditions. In this paper, we specifically focus on night-time and rainy conditions, which constitute a major challenge for AVs in practice. 

\subsection{Perception in Adverse Weather Conditions}

As discussed in Section \ref{3dod}, existing perception approaches have severe limitations in adverse weather conditions. Bijelic et al. present a multimodal dataset and a sensor fusion architecture for use in fog conditions \cite{bijelic2020seeing}. Jin et al. have introduced a rich annotated image dataset for rainy street scenes \cite{jin2021raidar}. ReViewNet \cite{mehra2020reviewnet} designs a network for safe autonomous driving in hazy weather conditions. However, these studies tend to focus on datasets that are curated for their respective studies and the generalization of performance to bigger, popular datasets such as KITTI \cite{deschaud2021kitti} and NuScenes \cite{nuscenes} is not well understood. In this paper, we focus on a generalized 3D-object detection pipeline that can handle night-time and rainy conditions using lidar and camera while also significantly improving performance at night on the NuScenes dataset. 

\subsection{AV Simulators}

Testing AV technology in real-life settings, particularly under poor weather conditions, is very expensive in terms of time, cost and resources, besides being very uncontrollable and not reproducible. Hence, AV simulators are attractive for testing these scenarios as a precursor to real-life testing. Many simulators have been proposed in the literature for simulating AV systems. These include CARLA \cite{dosovitskiy2017carla}, GTA V \cite{GTAV}, AirSim \cite{AirSim}, SUMO \cite{SUMO} and TORCS \cite{TORCS}. These simulators play a key role in generating multi-modal synthetic sensor data inputs for training machine learning networks with data augmentation. In comparison to other existing AV simulators, {CARLA} can model all three primary types of AV sensors, namely RGB cameras, lidars and radars. It is also open-sourced and well-documented, which provides flexibility in terms of generating different types of data with annotations such as semantic segmentation, instance segmentation and 3D bounding boxes. Many studies like KITTI-CARLA \cite{deschaud2021kitti} use CARLA to generate synthetic data for enriching existing datasets and to perform co-simulation with AV software stacks \cite{sural_cosim2023}. We use CARLA to complement the NuScenes dataset with synthetic data and to make up for the daytime and clear weather data bias in NuScenes. We also use this enriched data set to evaluate our \textit{ContextualFusion} approach in a variety of poor weather and lighting conditions.

\section{Our Methodology}
\label{sec:methodology}
In this section, we first describe the datasets used in our experiments. We then present the architecture of our multi-sensor fusion model named {\textit{ContextualFusion}} for 3D object detection.
\subsection{Datasets}
Our focus is on a fusion approach using camera and lidar data. We use the real-world NuScenes \cite{nuscenes} dataset along with our synthetic multimodal dataset named \textit{AdverseOp3D}.
\subsubsection{NuScenes}
NuScenes \cite{nuscenes} is a multimodal dataset collected using a vehicle that records various sensor data including a 360-degree high-resolution lidar and six cameras. Overall, there are about 1000 scenes in the dataset and, among these, each driving scene has around 40 annotated frames resulting in approximately 40,000 data samples, large enough for our study for camera and lidar fusion. Furthermore, each data sample is labeled with the contextual information of day-time, night-time and rain. Unfortunately, the dataset has only $~11.6\%$ of its samples during night-time, $~19.4\%$ of samples in rain and only $~1.6\%$ of the dataset is in rainy weather at night, one of the most challenging conditions for perception in AVs. Such bias is quite common in other AV 3D object detection datasets such as KITTI \cite{deschaud2021kitti} as well. This inherent bias in the dataset, in addition to poor image-based vision performance at night, leads to a significant deterioration of object detection accuracy in poor weather and lighting conditions. Our {\textit{ContextualFusion}} model tackles this degradation in performance under adverse environmental conditions.

\begin{figure}
\centering
\includegraphics[scale=0.16]{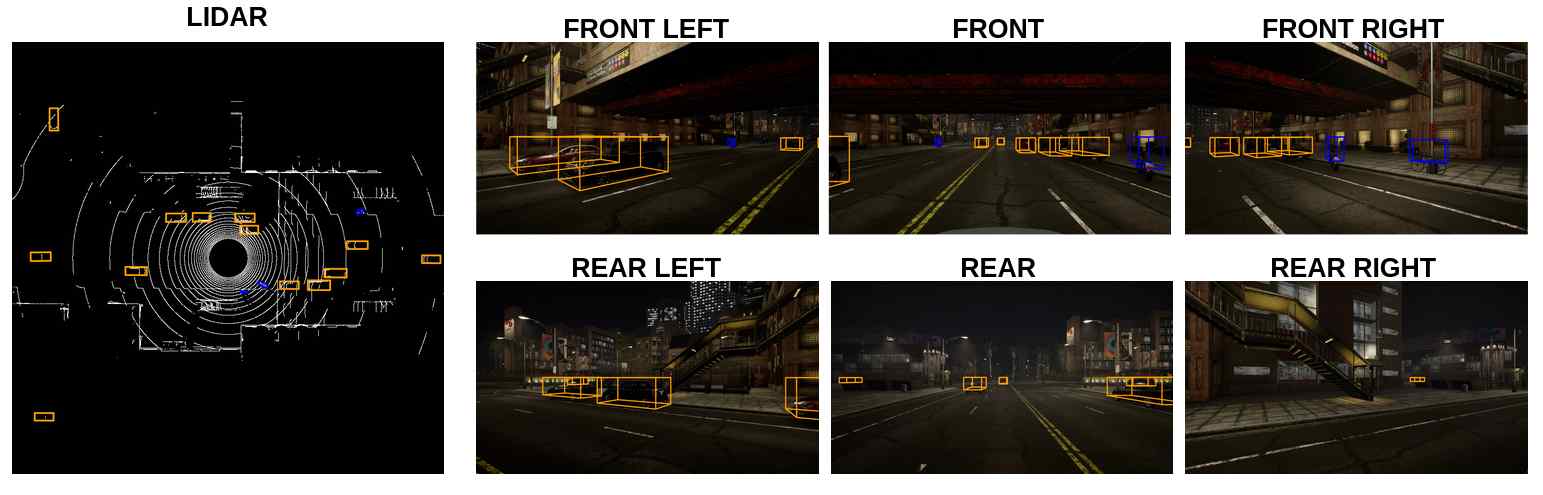}
\caption{Bounding box ground truth generated from CARLA in lidar and camera views corresponding to the NuScenes sensor configurations. The image brightness is enhanced for better visibility.} 
\label{bbox_gt}
\vspace{-3mm}
\end{figure}
\subsubsection{Synthetic Multimodal Dataset Creation}
We generate and utilize a synthetic adverse weather dataset in the NuScenes format named \textit{AdverseOp3D}. 
%Our goal is to augment the NuScenes dataset, which is biased towards day-time conditions, with additional night-time data that has lidar and camera information, annotated with 3D bounding boxes. 
We use the CARLA simulator \cite{dosovitskiy2017carla}, mount multiple sensors on the simulated AV, drive the simulated AV around (autonomously if needed) in the virtual world and collect data. To ensure that the data is compatible with NuScenes, we set up six cameras and one lidar in the same positions and orientations as described in the NuScenes dataset generation process. 
%This configuration yields a 360-degree 32-beam lidar on top and six cameras facing front, front right, front left, rear center, rear left and rear right. 
We generate the synthetic dataset after synchronizing these sensors in CARLA. For every point in the dataset, we command CARLA to generate a list of ground-truth bounding-box annotations in the 3D simulation environment. We then filter this list to ensure that only a relevant subset of these boxes is actually used, with relevance based on whether there is at least one lidar point within the bounding box. Examples of bounding boxes projected on a sample lidar point cloud are shown in Figure \ref{bbox_gt}. In the image, the boxes corresponding to objects such as cars and bikes are marked in red. For every data point, we also store the environmental information about day- or night-time conditions in addition to the presence or absence of rain, which is readily available as ground truth from the simulator.

\begin{figure}
\centering
\includegraphics[scale=0.11]{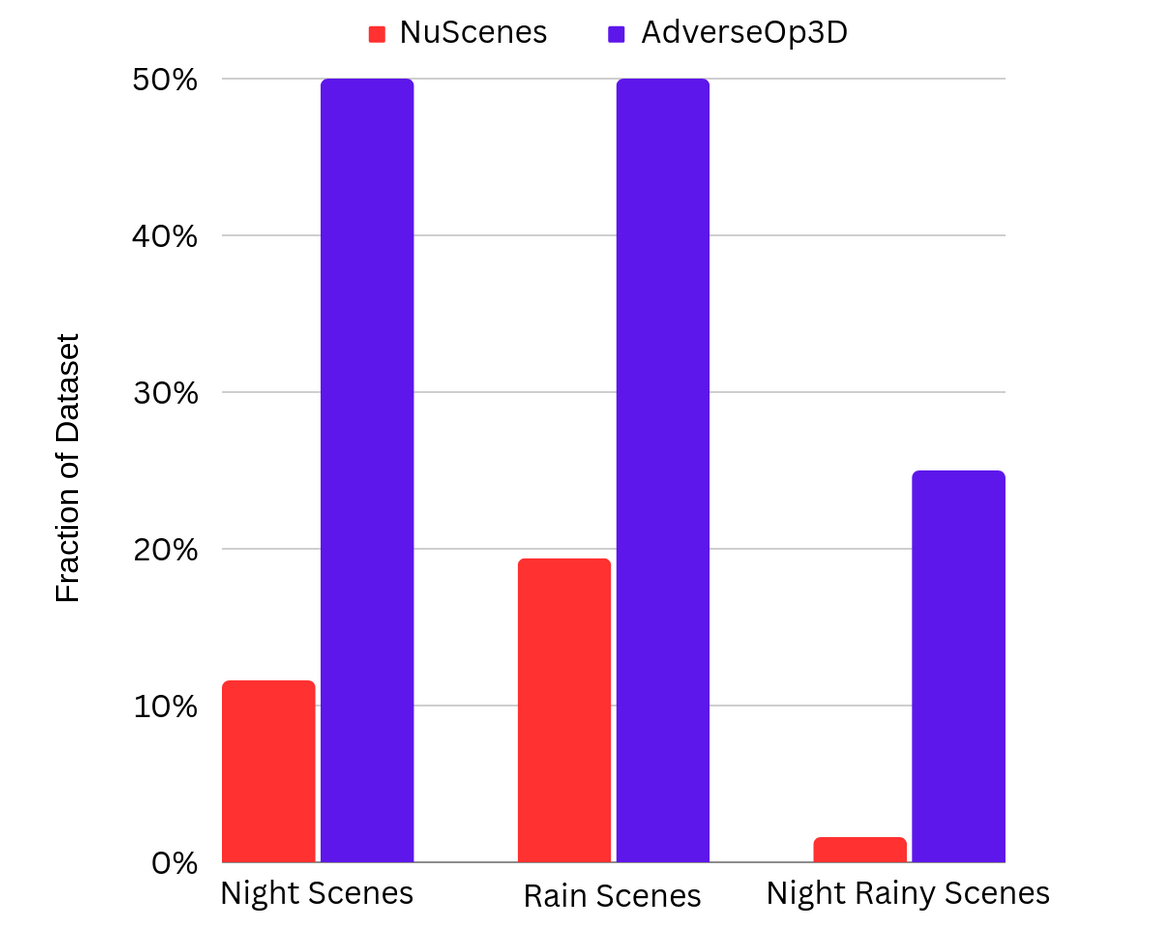}
\caption{The distribution of adverse operating conditions in our \textit{AdverseOp3D} dataset in comparison to the NuScenes dataset.} 
\label{adverseop3d_splits}
\vspace{-5mm}
\end{figure} 
CARLA simulates the effects of rain and night-time on images using post-processing. Specifically, the glare produced at night from streetlights, car taillights and reflections from wet roads get incorporated into the generated CARLA images, thereby enhancing the realism of generated data. Rain particles are also simulated in CARLA causing reasonable occlusion in images captured using cameras. Realism in the lidar data generated from CARLA is ensured using atmospheric attenuation models and intensity-based dropoff methods.

% Finally, a script takes in all of the data from CARLA, processes it and creates multiple JSON files that are linked by tokens in the NuScenes format. These files contain meaningful information about the ego pose, annotations of objects, sensor calibration and other associated attributes. The generated JSON files are appended to the original NuScenes JSON files to obtain the augmented dataset.
\begin{figure*}
\centering
\includegraphics[scale=0.2]{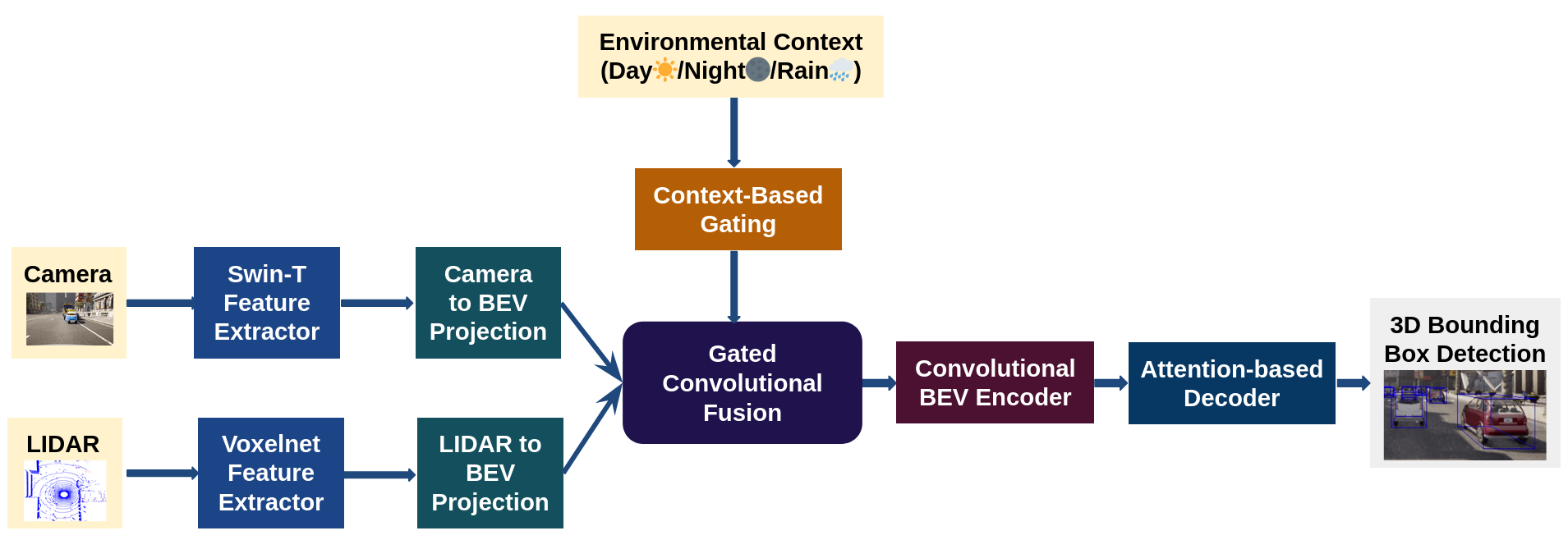}
\caption{Our \textit{ContextualFusion} Model Architecture} 
\label{architecture}
\end{figure*} 
Using the procedure described, we generate a large dataset across day, night, clear and rainy weather conditions called \textit{AdverseOp3D}. \textit{AdverseOp3D} contains more than 4000 samples with more than $75\%$ of the dataset being in adverse conditions (Night+Clear - $25\%$, Day+Rain- $25\%$ and Night+Rain - $25\%$). Figure \ref{adverseop3d_splits} shows the data distribution under adverse conditions in our synthetic \textit{AdverseOp3D} dataset compared against the imbalance in NuScenes. Countering the bias in NuScenes is a positive step towards safe all-weather autonomous driving. To evaluate \textit{ContextualFusion}, we augment the \textit{AdverseOp3D} dataset with a small subset of the NuScenes dataset consisting of 400 samples from 10 scenes ($1\%$ of the whole dataset). This is done to ensure better real-world generalizability of the model trained on CARLA data. While our high-level goal is to ensure that our method eventually improves object detection performance in adverse conditions in the real world, \textit{AdverseOp3D} also serves as a baseline for our evaluation and motivates the creation of more balanced datasets from the real world.  
\subsection{Architecture for \textit{ContextualFusion}}
% \subsubsection{Night NuScenes Dataset}
% [Shounak+Nishad]
% Dataset name - Night NuScenes
% (For eval)

Our \textit{ContextualFusion} model leverages domain knowledge about the physics of lidar and camera sensors. lidar, being an active sensor that does not rely on external light sources, generates a point cloud of its surroundings with similar effectiveness in day and night. A camera, on the other hand, relies on the presence of light sources such as the sun, streetlights or vehicle headlights. However, during heavy rain, lidar beams can be scattered by interfering rain droplets resulting in false obstacle detections. A camera image also faces occlusion problems when it is raining heavily or pitch dark, but its limitations do not necessarily overlap with that of a lidar. 

\textit{ContextualFusion}  incorporates domain knowledge about sensor types into a mid-level sensor fusion model utilizing the Bird's Eye View (BEV) space, while exploiting contextual information. We describe the architecture of our model next.
%Current lidar-camera fusion models use deep-learning networks for feature extraction from lidar, camera data and other sensing modalities. The most effective networks use a mid-level fusion approach as opposed to early or late fusion techniques. In this approach to fusion, information from lidar and camera sources is fused after feature extraction and projection to a 2D or 3D plane corresponding to camera or lidar. 
\subsubsection{Overview of \textit{ContextualFusion}}
Figure \ref{architecture} presents the architecture of \textit{ContextualFusion} that incorporates environmental context into a 3D object detection pipeline. This architecture is inspired by BEVFusion \cite{bevfusion}, but our model overcomes the shortcomings of BEVFusion in terms of night-time performance. Our \textit{ContextualFusion} network is built using the MMDetection3D \cite{mmdet3d} library based on PyTorch.
 
\subsubsection{3D Object Detection}
BEV space is effective for 3D object detection  since it is not geometrically lossy like what is obtained with projection from lidar to camera space where the depth information is lost, and points in 3D space that are highly separated in depth can be clubbed together in a camera image. Additionally, mapping camera features to lidar points suffers from the loss of semantic information for pixels that are not close to any lidar point, which is problematic in the presence of rain. In our setting, a common BEV space in which all camera and lidar features are projected is also useful.

In Figure \ref{architecture}, the camera feature extraction
block uses the SwinTransformer \cite{swint}, while the lidar feature extraction block uses the VoxelNet \cite{voxelnet} architecture inspired by BEVFusion. We use nine lidar sweeps from the NuScenes dataset to ensure that point clouds get accumulated over time and result in a dense version. For our synthetic \textit{AdverseOp3D} dataset, we do not need to use multiple sweeps since the generated synthetic data is already much more accurate and dense. Once
the features have been extracted, the camera features are transformed to the BEV space based on a
soft depth-estimation process that extracts features at various
depths based on images from the six cameras. 
%This can be computationally expensive \footnote{Details on speeding up this process are provided by Liu et al. \cite{bevfusion}}. 
For the lidar features, flattening is done along the z-axis to obtain the BEV
representation in the lidar space. Once both camera and lidar features are extracted, they are fed
into an environmental-context-based gating module, which in turn performs a gated convolutional fusion operation for fusing the two modalities. This is followed by an additional convolutional BEV encoder
that corrects local misalignments arising from sensor calibration errors.

The output head of our \textit{ContextualFusion} network is inspired by the Transfusion model \cite{transfusion}. A multi-head attention-based decoder is used, followed by 1-dimensional convolutional layers to predict bounding boxes and classes of interest. Specifically, the network predicts the center location of objects using a class-based center heatmap prediction unit along with regression heads that predict the dimension and orientation of these objects. The outputs of our network are 3D bounding boxes for objects
of interest such as cars, buses, pedestrians and bicycles. We support ten such classes for the NuScenes dataset currently. For each bounding box, the network predicts its center, its extent
and a yaw angle for its orientation.
%WRite about loss function from the model arch - multi-headed attention, position embedding,  etc

\subsubsection{Gated Convolutional Fusion}\label{gatedconv}
\begin{figure}
\centering
\includegraphics[scale=0.06]{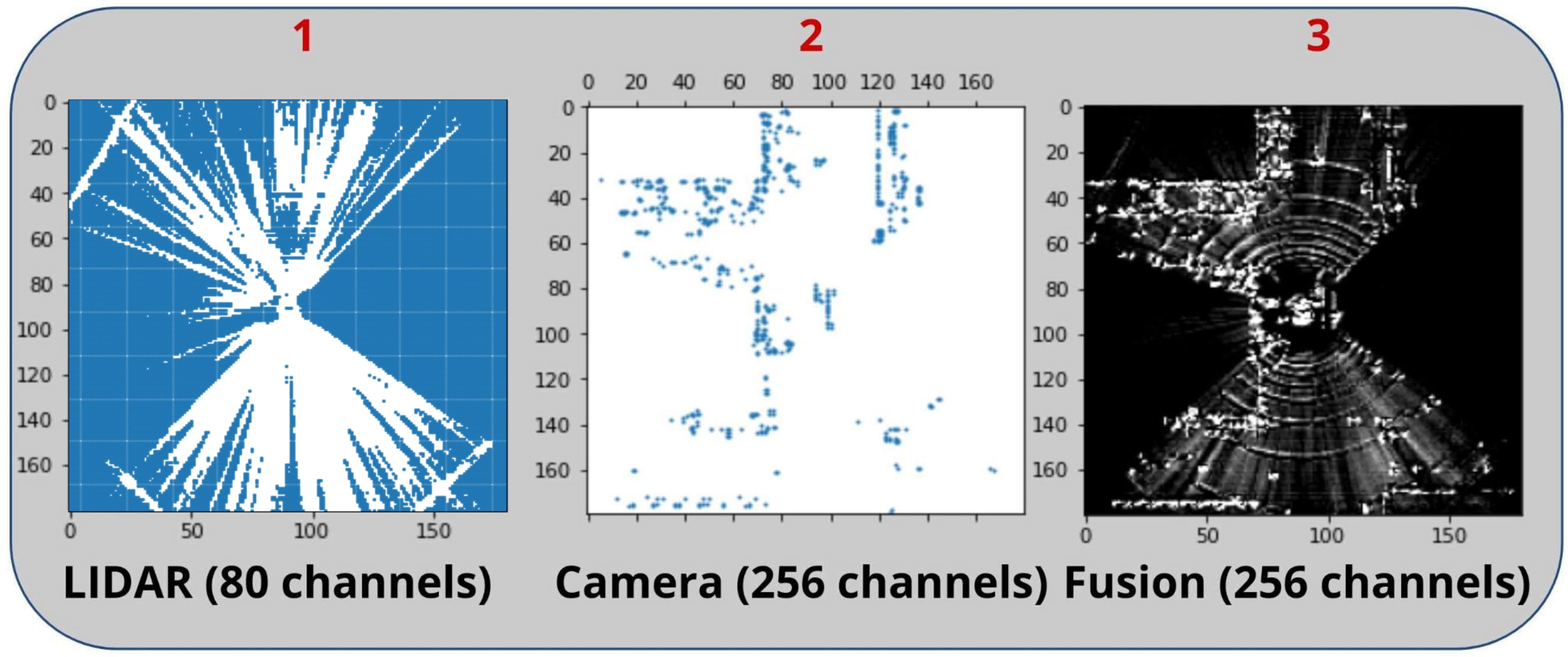}
\caption{Visualization of lidar and camera features before and after fusion} 
\label{feat_viz}
\vspace{-5mm}
\end{figure}

Our goal is to find an efficient technique to incorporate information about rain and nighttime conditions into the fusion block of the proposed network. To understand the feasibility of relative weighting of lidar and camera in the shared BEV space based on context, the BEV features corresponding to before and after fusion can be seen in Figure \ref{feat_viz}. These correspond to the scenario with lidar data and associated bounding box annotations seen in Figure \ref{bbox_gt}. 

The intermediate feature representation in the BEV from the lidar features has 80 channels of size $180\times180$, and the equivalent representation from the camera features has 256 channels, also of size $180\times180$. A sparse image of one lidar channel is visualized in Figure \ref{feat_viz} along with a sparse image of the camera channel. These represent features that mark the edges of the surrounding world in BEV. These are concatenated and fed into a convolutional network to create an output feature representation of 256 channels of size $180\times180$. Since the same environment is visualized across both lidar and camera features with a 1:1 pixel level correspondence in Figure \ref{bbox_gt}, a relative lidar-camera weighting approach can be used in our model.  

Since our fusion approach is based on convolution with a weighted average of all lidar and camera channels, the lidar and camera features are treated equally in the input to the convolution block. Since the sensory response of lidars and cameras varies across environmental conditions, the operating environmental context is used to create a context-based gating block. Specifically, we use binary values for rain and night information in every sample as the context input to this network. The output from this block with the context information is used to weigh the inputs of the gated convolution block. For instance, at night, the lidar data will be given more weightage compared to the camera, with the relative weights learnt with a linear layer that is added before convolution.

We now define Gated Convolutional Fusion (\textit{GatedConv}), with the valid assumption that multiple modalities are already in a shared representation space with the same height and width for each channel, i.e., BEV in our case. The operation is defined with the following notation:

\begin{itemize}[noitemsep]
    \item There are $C_1$ channels for Sensing Modality 1 and $C_2$ channels for Sensing Modality 2.
    \item Based on the context-based gating module, Sensing Modalities 1 and 2 have context-weight vectors $G_1$ and $G_2$.
    \item The output of gated convolution has $C_{out}$ channels.
\end{itemize} 

We choose a stride and padding of one in the convolution operation to ensure that the output feature dimension within each channel is the same as that of the input and has one-to-one correspondence at a pixel level as in Figure \ref{feat_viz}.

The output of GatedConv is defined as:
\begin{gather*}
GatedConv(C_{out_i})=bias(C_{out_i}) \\ + \sum_{j=0}^{C_1-1}G_1(j)*weight(C_{out_i},j)*input(j) \\ +\sum_{k=0}^{C_2-1}G_2(k)*weight(C_{out_i},k)*input(k) \\
\end{gather*}

Here, unlike a regular convolution across $C_1+C_2$ channels, $C_1$ channels are gated by a vector $G_1$ corresponding to Sensing Modality 1, and $C_2$ channels are gated by a vector $G_2$ corresponding to Sensing Modality 2. We consider two variations of this approach in our experiments: (i) the sensing modality-specific weights $G_1$ and $G_2$ vary independently across the channels in each modality, and (ii) $G_1$ and $G_2$ are restricted to be uniform across sensing modalities. We refer to these two methods as \textit{ContextualFusion (Independent)} and \textit{ContextualFusion (Constrained)} respectively. Independence in these weights allows for the possibility of varied weightage to different channels of the network within each sensing modality based on the context. This is a more general approach with fewer assumptions and is expected to work better on a complex dataset with multiple classes of interest. However, the constrained approach is a simpler model and is therefore, worth investigating. We compare the effectiveness of both these approaches in Section \ref{sec:experiments}. 

% \begin{table*}
%     \centering
%     \caption{\textit{ContextualFusion} on NuScenes- the AP scores (in $\%$) for individual classes are reported for some classes of interest along with the overall mAP for 10 classes}
%     \vspace{2mm}
%     \begin{tabular}{l|c|c|c|c|c|c|c|c|c|c|c} \toprule
%         {} & {Car} & {Bus}& {Construction Vehicle}& {Pedestrian}& {Motorcycle}& {Bicycle}& {Traffic Cone}& {mAP} \\ \midrule
%         \textbf{BEVFusion} & 88.8 & 73.2 & 28.3 & \textbf{88.2} & 73.6 & 59.3 & \textbf{79.7} & 66.84\\
%         \textit{ContextualFusion} & \textbf{88.8} & \textbf{75.7} & \textbf{29.2} & 87.7 & \textbf{76.6} & \textbf{62.5} & 79.3 & \textbf{67.75}\\ \bottomrule
%     \end{tabular}
%     \label{nusc_results}
% \end{table*}
\begin{table*}
    \centering
    \vspace{2mm}
    \caption{Our approach \textit{ContextualFusion (CF)} evaluated on NuScenes data along with rainy and night-time subsets of the dataset. AP scores (in $\%$) for individual classes are shown along with the overall mAP. '-' refers to classes without samples in the Night NuScenes validation set and C and L refer to camera and lidar modalities respectively. *C.V. represents construction vehicles}
    \begin{tabular}{l|c|c|c|c|c|c|c|c|c|c|c|c} \toprule
        {Method} & {Modality} & {Car} & {Truck} & {Bus} & {Trailer} & {C.V.*} & {Pedestrian}& {Motorcycle}& {Bicycle}& {Traffic Cone} & {Barrier} & {mAP} \\ \midrule
        \multicolumn{12}{c}{\multirow{2}{*}{\hspace*{2cm}\textbf{FULL NUSCENES}}} \\
        \multicolumn{12}{c}{}\\
        %\
        \hline
        \textbf{BEVDet \cite{bevdet}} & C & 53.8 & 29.0 & 40.4 & 17.7 & 9.2 & 38.6 & 34.1 & 26.4 & 54.4 & 51.7 & 35.5\\
        \textbf{Transfusion-L \cite{transfusion}} & L & 86.9 & 61.0 & 72.6 & 42.1 & 27.1 & 86.5 & 71.8 & 56.2 & 73.3 & 69.7 & 64.7\\
        \textbf{BEVFusion \cite{bevfusion}} &  C+L & 88.5 & 62.1 & 72.7 & \textbf{43.6} & 28.4 & 87.4 & 75.7 & 58.5 & 79.3 & 71.0 & 66.8\\
        \textbf{CF \textit{(Constrained)}} & C+L & 88.7 & \textbf{64.2} & \textbf{75.3} & 43.3 & \textbf{28.8} & 87.4 & \textbf{76.1} & \textbf{61.6} & 79.0 & \textbf{72.0} & \textbf{67.5}\\ 
        \textbf{CF \textit{(Independent)}} & C+L & \textbf{89.2} & 62.4 & 73.4 & 43.5 & 28.8 & \textbf{88.4} & 76.0 & 59.3 & \textbf{79.6} & 71.5 & 67.2\\ \bottomrule
        \multicolumn{12}{c}{\multirow{2}{*}{\hspace*{2cm}\textbf{RAIN NUSCENES}}} \\
        \multicolumn{12}{c}{}\\
        \hline
        \textbf{BEVDet \cite{bevdet}} & C & 53.2 & 33.2 & 50.0 & 14.9 & 11.8 & 26.7 & 21.7 & 26.9 & 51.7 & 66.1 & 35.6\\
        \textbf{Transfusion-L \cite{transfusion}} & L & 88.4 & 62.0 & 78.5 & 44.1 & 22.3 & 78.8 & 74.6 & 43.8 & 69.5 & 82.5 & 64.5\\
        \textbf{BEVFusion \cite{bevfusion}} & C+L & 90.1 & 64.8 & 80.3 & \textbf{42.3} & \textbf{25.4} & 78.6 & 79.1 & 55.6 & 77.6 & 81.4 & 67.5\\
        \textbf{CF \textit{(Constrained)}}& C+L & 90.1 & 64.9 & 80.3 & 42.2 & 25.3 & 78.6 & 79.1 & 55.3 & 77.6 & 81.4 & 67.5\\
        \textbf{CF \textit{(Independent)}}& C+L & \textbf{90.8} & \textbf{66.1} & \textbf{81.9} & 41.2 & 24.6 & \textbf{81.7} & \textbf{81.9} & \textbf{58.5} & \textbf{78.9} & \textbf{82.9} & \textbf{68.9}\\ \bottomrule
        \multicolumn{12}{c}{\multirow{2}{*}{\hspace*{2cm}\textbf{NIGHT NUSCENES}}} \\
        \multicolumn{12}{c}{}\\
        \hline
        \textbf{BEVDet \cite{bevdet}} & C & 48.1 & 20.3 & - & - & - & 22.6 & 15.6 & 12.2 & - & 34.6 & 15.3\\
        \textbf{Transfusion-L \cite{transfusion}} & L & 88.1 & 81.9 & - & - & - & 44.0 & 78.5 & 31.5 & - & 40.5 & 36.7\\
        \textbf{BEVFusion \cite{bevfusion}} & C+L & 83.4 & 65.3 & - & - & - & 55.3 & {66.2} & 30.4 & {-} & {67.0} & 61.3\\
        \textbf{CF \textit{(Constrained)}} & C+L& 89.3 & \textbf{82.9} & - & - & - & \textbf{57.4} & \textbf{79.7} & {44.4}& {-} & {73.2} & {71.1}\\
        \textbf{CF \textit{(Independent)}} & C+L& \textbf{89.9} & 80.5 & - & - & - & 51.4 & {78.0} & \textbf{55.8} & {-} & \textbf{82.5} & \textbf{73.0}\\ \bottomrule
    \end{tabular}
    \label{night_results}
\end{table*}
\subsection{Obtaining Context Information in an AV} \label{viability_context}
We use the information about the presence of day/night and rain/clear weather conditions as an input to the deepnet based on annotations provided in the NuScenes dataset. For the CARLA-based dataset, we obtain weather information directly from the simulator. While contextual information is not necessarily directly available on an AV, there are viable sources. Information such as day/night or rain/non-rainy weather conditions can be obtained from sources offering or using weather information. In a modern car, automatic windshield wipers sense rain with a sensor and can adjust their operation accordingly. The same information can be tapped into by an AV and used to detect rainy conditions. For day/night-time conditions, automated headlight sensors can sense the amount of ambient light in the environment and correspondingly turn headlights off or on. This information can also be accessed to provide context information for use in \textit{ContextualFusion}. In addition, if the AV is connected to the Internet, (potentially delayed) weather information can be obtained and used. Alternatively, a CNN-based architecture such as WeatherNet \cite{weathernet} can be used to process camera images and determine the weather context.

\section{Evaluation of \textit{ContextualFusion}}\label{sec:experiments}
In this section, we conduct experiments to evaluate the effectiveness of our methods. For quantitative evaluation, we use Average Precision (AP) scores for each class of interest to determine their 3D object detection performance. AP is defined as the area under the precision-recall curve for 3D bounding box predictions and is expressed as a percentage. We also use mAP (mean AP) over all classes to get an overall performance estimate and compare with existing approaches.

\subsection{Quantitative Results}
We conduct two sets of experiments for incorporating context in the contextual gating module as introduced in Section \ref{gatedconv}. In the first set of experiments, we use the real-world NuScenes dataset in addition to its rain and night subsets to evaluate our approaches. The training set has around 34,000 samples in NuScenes, while the validation set has about 6,000 samples. The night-time split for validation has 600 samples and the rainy validation split has around 1,080 samples. The results for the \textit{ContextualFusion (Constrained)} and \textit{ContextualFusion (Independent)} approaches discussed in Section \ref{gatedconv} are summarized in Table \ref{night_results} along with a comparison with several state-of-the-art methods. The results are shown for 10 classes of interest that are part of the NuScenes dataset. All of the results are from the NuScenes validation split. We choose this subset because the NuScenes leaderboard-based evaluation for the test split does not have annotations for night or rain available as part of the input during testing. As discussed earlier in Section \ref{viability_context}, obtaining these inputs is feasible for model deployment on an AV and our usage is therefore justified. 

We trained the BEVFusion model using the training schedule and the pretrained lidar and camera backbones as described in \cite{bevfusion} on the entire NuScenes dataset. This trained model is used as our baseline for transfer learning to our \textit{ContextualFusion} model. Transfer learning helps generalize knowledge from an existing model onto a new domain without re-training the original model extensively. In our case, the additional trainable weights in \textit{ContextualFusion} which perform context-based gating are trained, while pre-loading other weights from the BEVFusion pipeline we trained on Nvidia A6000 GPUs.
\begin{table}
    \centering
    \caption{\textit{ContextualFusion} (CF) Performance on AdverseOp3D + NuScenes mini. AP and mAP scores are reported in $\%$}
    \begin{tabular}{l|c|c|c} \toprule
        \multicolumn{1}{p{1.6cm}}{} & \multicolumn{1}{p{1.6cm}}{\centering BEVFusion\cite{bevfusion}}& \multicolumn{1}{p{1.8cm}}{\centering \textit{ CF      \hspace{1cm}(Constrained)}}& \multicolumn{1}{p{1.8cm}}{\centering\textit{ CF \hspace{1cm}(Independent)}} \\ \midrule
        \textbf{Car AP}  & {74.0} & \textbf{78.7} & {{78.5}} \\ 
        \textbf{Pedestrian AP}  & {63.9} & {66.5} &  {\textbf{67.6}} \\ 
        \textbf{Truck AP} & {24.6} & {28.5} & {\textbf{32.1}} \\ 
        \textbf{Bus AP} & 52.7 &  {57.1} & {\textbf{67.6}} \\ 
        \textbf{Overall mAP} & {53.8}  & {57.7} &{\textbf{60.0}} \\ \bottomrule
    \end{tabular}
    \label{synthetic_table}
\vspace{-5mm}
\end{table}
Table \ref{night_results} shows the evaluation of our methods on the full NuScenes dataset followed by subsections for its rain and night subsets. As can be seen, our \textit{ContextualFusion (CF)} approach works slightly better than the best-performing BEVFusion method on the entire dataset. Moreover, CF performs much better than other existing approaches for 3D object detection that use only lidar or camera for detection. In the rainy subset of NuScenes, \textit{ContextualFusion (Independent)} beats the baseline with an improvement of $1.4\%$. The usefulness of CF is much more pronounced in the night-time dataset from NuScenes where the  \textit{CF (Independent)} approach outperforms BEVFusion by $11.7\%$ in terms of overall mAP and even the  \textit{CF (Constrained)} approach improves performance by nearly $10\%$. While both the independent and constrained approaches work well, differential importance across features from varying heights in case of \textit{CF(Independent)} performs better at night. A key point to be noted is that, for the night-time validation dataset, the bus, trailer, construction vehicle and traffic cone classes are not present in the data. Hence, the overall mAP is calculated based only on the other six classes that are present. 

The drop in performance at night-time for the BEVFusion network is much more pronounced for some classes corresponding to smaller objects such as bicycles and motorcycles. Our network compensates for these issues at night with the context-based gating network, resulting in over $25\%$ improvement in mAP for the case of bicycles. With \textit{ContextualFusion}, the performance for several classes of interest at night are nearly as good as the performance in daytime. This positive result is obtained because lidar gets trained to be given a higher importance at night than the camera which ensures better pickup of the structure of objects, when camera images are not clearly visible in low-lighting conditions.

Motivated by the success of \textit{ContextualFusion} on the rainy and night-time subsets of NuScenes, in the next set of experiments, we compare our approach with BEVFusion on our synthetic dataset which is balanced across various environmental conditions.
%The subset of NuScenes adds some real data to both the training and validation sets in our experiments. This is followed by a performance assessment on the entire NuScenes dataset. We also test on the Night NuScenes and the Rain NuScenes datasets, which contain only the night-time and rainy scenes from NuScenes respectively.
\begin{figure}
\centering
\includegraphics[scale=0.065]{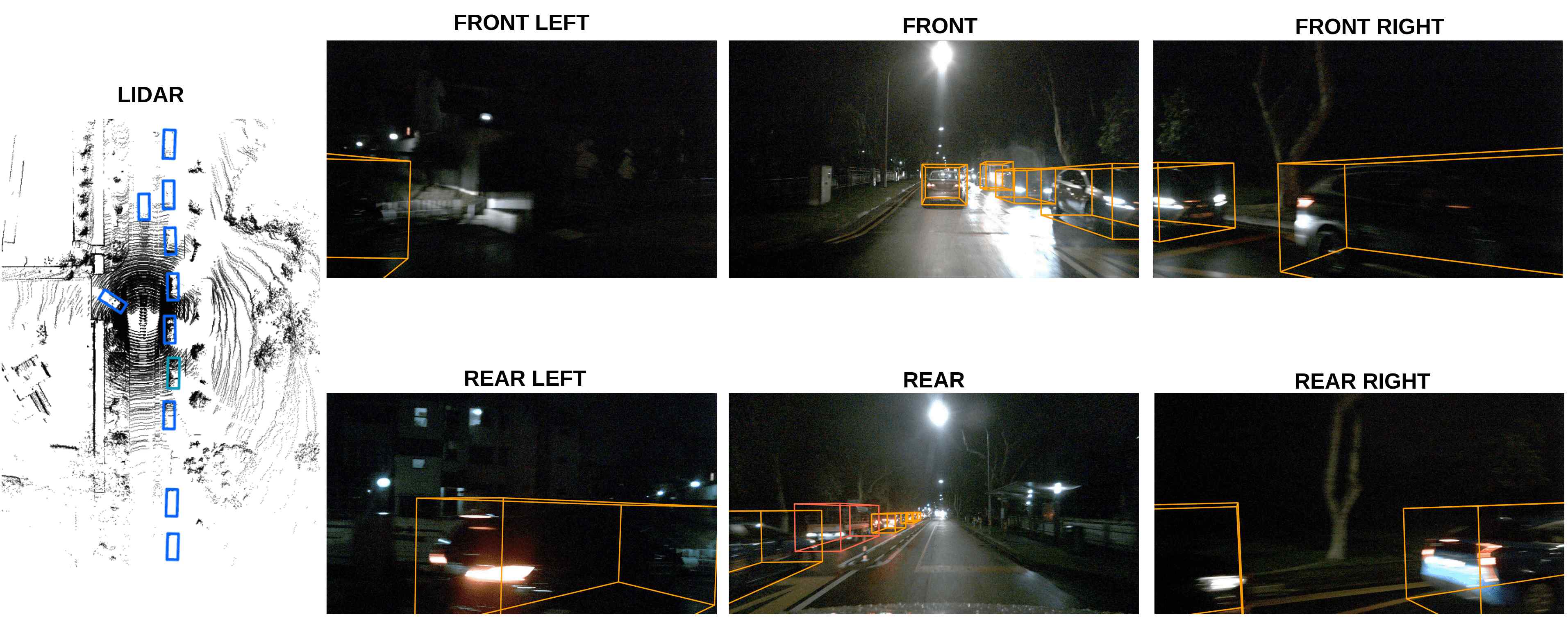}
\caption{3D object detection performance of \textit{ContextualFusion} on NuScenes night-time data} 
\label{image:nusc_night}
\end{figure} 
We specifically use a combination of our \textit{AdverseOp3D} dataset with a small subset of NuScenes to both evaluate \textit{ContextualFusion} and test the practicality of our data generation method to create a balanced dataset. 

The evaluation results on this dataset are summarized in Table \ref{synthetic_table}. We only use four classes of interest (car, truck, bus and pedestrian) since not all of the NuScenes classes have enough samples for a reliable and large-scale evaluation in this smaller dataset. \textit{ContextualFusion (Independent)} gives the best mAP of $60.0\%$ as well as the best individual APs for all four classes of interest. Additionally, we note that the \textit{AdverseOp3D } dataset has a 50-50 split between day and night scenes. Consequently, the \textit{ContextualFusion} approach works significantly better than BEVFusion on the entire dataset. Compared to BEVFusion, \textit{ContextualFusion} gets an improvement of $6.2\%$. As was expected, independently varying weights work better, but constrained weights across the modalities work fairly well too.
\begin{table}
    \centering
    \caption{Inference speed comparison on the NuScenes validation set using a single A6000 GPU}
    \begin{tabular}{l|c} \toprule
        \multicolumn{1}{p{1.6cm}}{\centering Method}& \multicolumn{1}{p{1.8cm}}{\centering Inference Speed (FPS)}\\ \midrule
        \textbf{PointAugmenting\cite{wang2021pointaugmenting}}  & {2.8} \\ 
        \textbf{Transfusion\cite{transfusion}}  & {5.5} \\ 
        \textbf{DeepInteraction\cite{deepinteraction}} & {3.1} \\ 
        \textbf{BEVFusion\cite{bevfusion}} & {\textbf{9.8}} \\ 
        \textbf{CF (\textit{Constrained})-Ours} & {9.7} \\
        \textbf{CF (\textit{Independent})-Ours} & {9.7}\\ \bottomrule
    \end{tabular}
    \label{inference_times}
\end{table}

We next evaluate the inference time of our model on a single A6000 GPU. A comparison of our results with other existing camera+lidar models is shown in Table \ref{inference_times}. The inference speed for {\textit{ContextualFusion}} is 9.7fps, which is similar to BEVFusion, while outperforming other methods significantly. 

\subsection{Visualization of Performance and Discussion}

In this subsection, we visualize and discuss the results obtained by \textit{ContextualFusion}. Figure \ref{image:nusc_night} shows the performance of our model at night on NuScenes data. The lidar image shows objects identified in the BEV frame using blue and green boxes. The same objects are also visualized with respect to six camera images. Observing the camera images at night, due to low visibility and glare from vehicle headlights and streetlights, the cameras are not very effective for accurate estimation. lidar should and does play a key role in these scenarios, and this property is exploited by \textit{ContextualFusion}. 
% \begin{figure}
% \centering
% \includegraphics[scale=0.07]{comparison_1img.jpg}
% \caption{Comparison of ground truth (left) vs BEVFusion predictions (middle) vs \textit{ContextualFusion} predictions (right) on night-time NuScenes data} 
% \label{img:comp_1img}
% \end{figure} 
% Figure \ref{img:comp_1img} shows an instance similar to the one described in Section \ref{introduction}. However, one of the cars in this scene further to the back is partially occluded by a car immediately behind the ego vehicle, making detection more challenging. BEVFusion has a high reliance on camera modality even at night which fails to detect this vehicle, but \textit{ContextualFusion} detects the vehicle. This distinction can be visualized in both the lidar and back camera views. 
\begin{figure}
\centering
\includegraphics[scale=0.07]{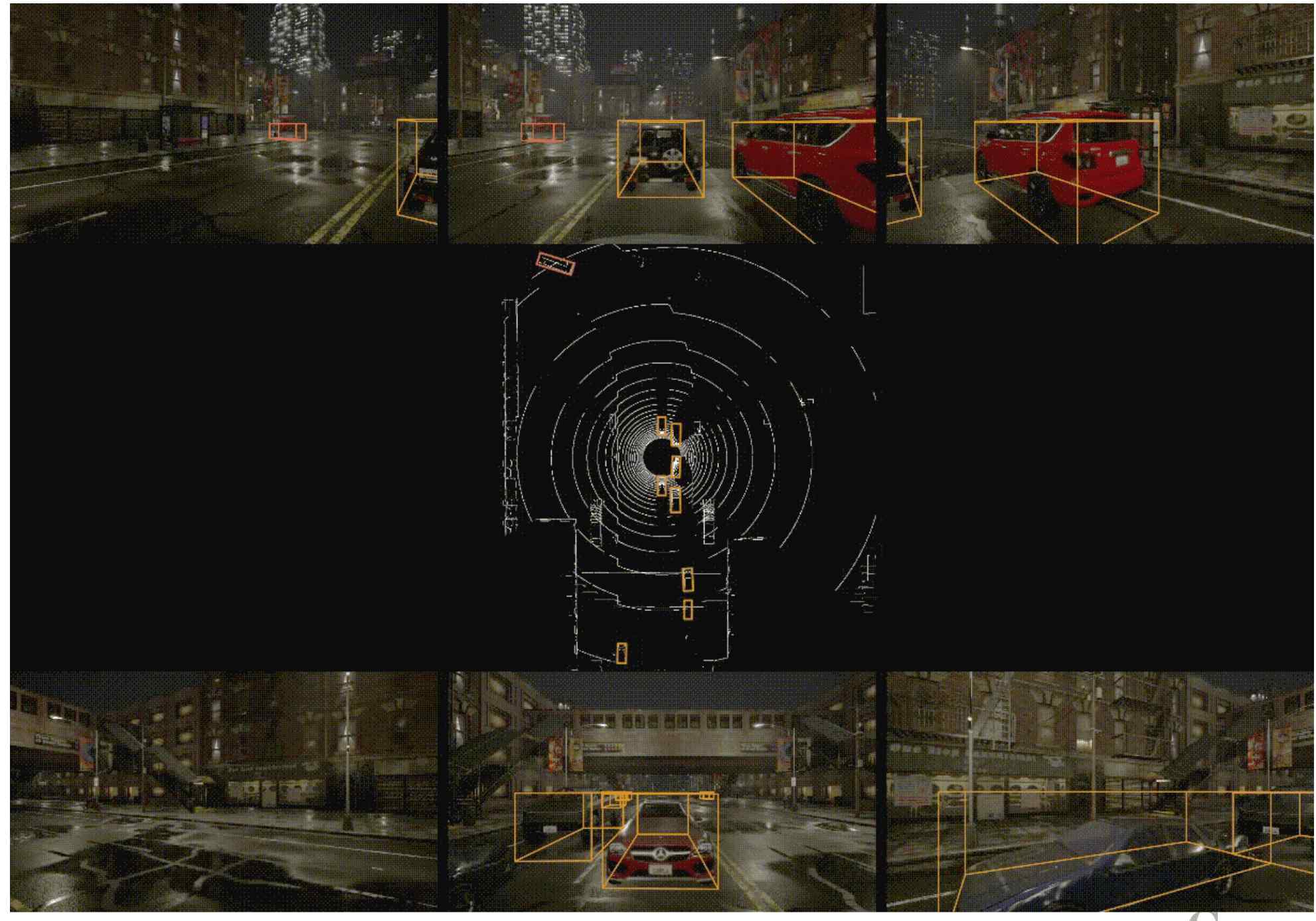}
\caption{3D object detection by \textit{ContextualFusion} on AdverseOp3D (CARLA generated) data at night while raining} 
\label{rain_night}
\vspace{-5mm}
\end{figure} 
% \begin{figure}
% \centering
% \includegraphics[scale=0.12]{real_result.jpg}
% \caption{3D object detection by our model on NuScenes daytime data} 
% \label{real_result}
% \end{figure} 
%a comparison of the ground truth versus the prediction done by the model on CARLA daytime data. The top three camera images within each set of images correspond to the front left, front and front right cameras respectively. Similarly, the bottom three correspond to the rear left, rear center and rear right cameras respectively. The image in the middle is the lidar data in BEV. This figure illustrates that our network performs quite well on real-world data and its predictions are comparable to the ground truth annotations. Noticeably, the objects of interest that are seen in the camera images are also visible in the lidar images.
Figure \ref{rain_night} shows that \textit{ContextualFusion} detects objects very well under adverse weather conditions such as rain at night-time in the \textit{AdverseOp3D} dataset. The brightness is enhanced in the image of Figure \ref{rain_night} for better visualization. We also observe that, while CARLA generated data simulates reflection of lights in road puddles, glares from light sources and atmospheric interaction of lidar beams, there is a non-negligible gap between these distributions and real-world data. More intensive style-transfer techniques such as the approach by Richter et al. \cite{photorealism} may need to be used in the future to make CARLA-generated data match real-world images better, yielding a more realistic dataset. 

% The predictions done by \textit{ContextualFusion} on the daytime data from NuScenes are shown in Figure \ref{real_result}. The objects of interest are clearly identified, localized relatively well, showing that \textit{ContextualFusion} does not lead to a performance drop in daytime, while improving night-time performance significantly. This trend is also seen in Table \ref{night_results} under the ``Full NuScenes" category where most of the dataset is in daytime.

\begin{figure}
\centering
\includegraphics[scale=0.18]{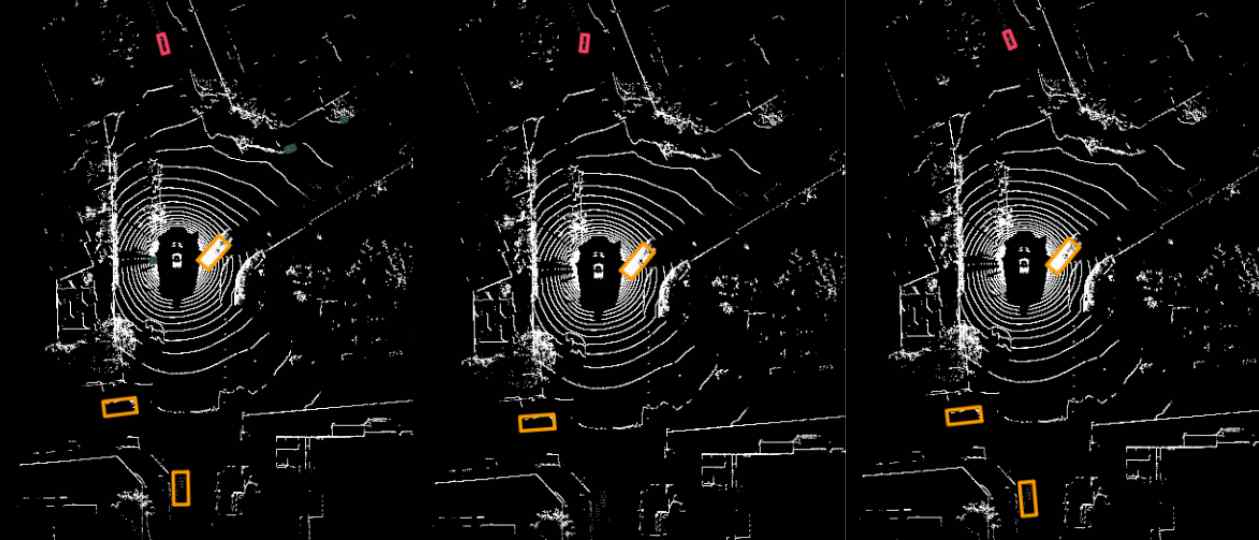}
\caption{Comparison of ground truth (left) vs BEVFusion predictions (middle) vs \textit{ContextualFusion} predictions (right) on night-time NuScenes data} 
\label{night_viz}
\vspace{-5mm}
\end{figure}

Figure \ref{night_viz} provides a BEV illustration of the performance of \textit{ContextualFusion} on the night-time NuScenes data. The leftmost image shows the ground truth annotations in the BEV for a night-time NuScenes scenario. The middle image shows the predictions of our baseline  BEVFusion on the same data point. It is noticeable that the car at the bottom in the ground truth (marked by a yellow box) is missing from the BEVFusion prediction. A motorcycle marked by a pink box also has a different orientation in the prediction compared to the ground truth. However, with \textit{ContextualFusion}, the motorcycle is in the correct orientation and the car at the bottom is detected as well. In summary, even visually, our approach picks up more meaningful objects of interest at night-time, which is indicative of the progress needed for safe AV driving in real-world environments and the usefulness of a context-based approach.
%Thus, visually we can see the model performs very well in synthetic adverse weather conditions and also in real-world data in spite of the fact that the dataset on which it was trained has very few real-world data points.  
\section{Concluding Remarks}
\label{sec:conclusion}
Autonomous vehicles must be able to drive safely under adverse operating conditions including poor lighting and weather conditions. We have designed an environmental context-based fusion network called
\textit{ContextualFusion} that adjusts the relative importance of
camera and lidar data depending on day- or night-time conditions and also based on rain. We introduce a Gated Convolutional Fusion (\textit{GatedConv}) operation that fuses features using context-based gating. We also created a synthetic dataset called \textit{AdverseOp3D}, which contains scenes from adverse weather conditions to counter existing bias towards good weather conditions. On the real-world NuScenes dataset at night-time, we obtain an mAP improvement of $11.7\%$ with the best-performing variant of our network
over state-of-the-art methods, while yielding a similar inference time. Additionally, \textit{ContextualFusion} produces an mAP improvement of $6.2\%$ on the entire \textit{AdverseOp3D} dataset compared to existing best-performing networks. Hence, \textit{ContextualFusion} significantly improves 3D object detection performance at night-time. 

The development of context detection frameworks and the subsequent usage of context-aware methods for perception and localization across all operational domains is an important direction for future AV research. In the future, our \textit{AdverseOp3D} dataset can also be further enhanced  with photorealism effects to reflect real-world distributions. 

\section*{Acknowledgement}
This work is funded by the US Department of Transportation under its ADS program. 

\printbibliography

\end{document}